\documentclass[conference]{IEEEtran}
\usepackage{comment}
\usepackage{amsmath}
\usepackage{amsfonts}
\usepackage{amssymb}
\usepackage{todonotes}
\usepackage{balance}
\usepackage{color}
\usepackage{graphicx}
\usepackage{array}
\usepackage{color,soul}
\usepackage{verbatim}
\usepackage{float}
\usepackage[caption=false]{subfig}

\usepackage[ruled,vlined,linesnumbered]{algorithm2e}
\title{Utilizing Priming to Identify Optimal Class Ordering to Alleviate Catastrophic Forgetting}

\author{\IEEEauthorblockN{Gabriel Mantione-Holmes}
\IEEEauthorblockA{\textit{Department of Computer Science} \\
\textit{Lewis \& Clark College} \\
Portland, OR \\
gabriel@lclark.edu}
\and
\IEEEauthorblockN{Justin Leo}
\IEEEauthorblockA{\textit{Department of Computer Science} \\
\textit{University of Colorado}\\
Colorado Springs, CO \\
jleo@uccs.edu}
\and
\IEEEauthorblockN{Jugal Kalita}
\IEEEauthorblockA{\textit{Department of Computer Science} \\
\textit{University of Colorado}\\
Colorado Springs, CO \\
jkalita@uccs.edu}
}

\begin{document}
    \maketitle

    \begin{abstract}
    In order for artificial neural networks to begin accurately mimicking biological ones, they must be able to adapt to new exigencies without forgetting what they have learned from previous training. Lifelong learning approaches to artificial neural networks attempt to strive towards this goal, yet have not progressed far enough to be realistically deployed for natural language processing tasks. The proverbial roadblock of catastrophic forgetting still gate-keeps researchers from an adequate lifelong learning model. While efforts are being made to quell catastrophic forgetting, there is a lack of research that looks into the importance of class ordering when training on new classes for incremental learning. This is surprising as the ordering of ``classes" that humans learn is heavily monitored and incredibly important. While heuristics to develop an ideal class order have been researched, this paper examines class ordering as it relates to priming as a scheme for incremental class learning. By examining the connections between various methods of priming found in humans and how those are mimicked yet remain unexplained in life-long machine learning, this paper provides a better understanding of the similarities between our biological systems and the synthetic systems while simultaneously improving current practices to combat catastrophic forgetting. Through the merging of psychological priming practices with class ordering, this paper is able to identify a generalizable method for class ordering in NLP incremental learning tasks that consistently outperforms random class ordering.
    \end{abstract}
    
    \begin{figure*}[h!]
      \includegraphics[width=\linewidth]{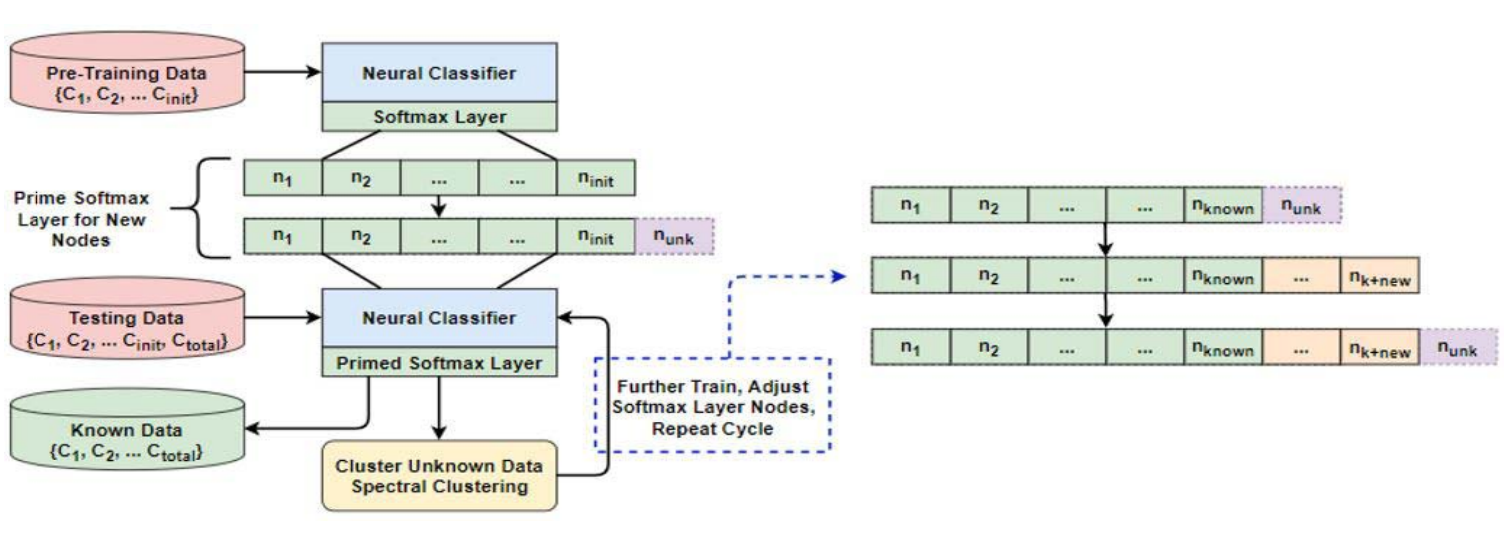}
      \caption{ A representation of the CCT method that adds a priming node at the classifiers softmax layer \cite{leo_incremental_2021}. The $n_{unk}$ represents the priming node that allows the model to gain additional nodes as classes are added.}
      \label{fig:CCT_Model}
    \end{figure*}
    
    \section{Introduction}
    Artificial neural networks have surpassed human abilities on a front of tasks. Human brains, unlike their synthetic counterparts, are hardly static in that they can, over their lifetime, learn new tasks while still retaining the ability to perform previously learned tasks.
    The same cannot be said about our current methods for isolated learning \cite{chen-liu-2016-lifelong}. Isolated learning, while proven useful, has real world limitations in that it cannot use previously learned knowledge to facilitate learning new tasks while still retaining information. When an isolated learning model is retrained on new data, it typically suffers from \emph{catastrophic forgetting}. Catastrophic forgetting, also known as \emph{catastrophic interference} \cite{MCCLOSKEY1989109} is the process by which a model loses the ability to classify data on which it has previously been trained \cite{li_total_2021}. 
    
    In order to perform well outside a controlled environment, machine learning models must be continually trained on batches of data from new classes while maintaining high accuracy in  previously trained classification tasks. This paradigm is variously called incremental learning, continual learning, sequential learning, or lifelong learning \cite{leo_survey_nodate}. Lifelong learning can benefit in ways that multi-class learning and isolated learning cannot. The major drawback shared by multi-class and isolated learning is that they assume data during training represent all the data and tasks that will be encountered in the real world. Lifelong learning, on the other hand, assumes that new tasks and data will naturally present themselves later. While lifelong learning models in computer vision have claimed a modicum of success, the mitigation of catastrophic forgetting among NLP models still has not met acceptable metrics \cite{greco-etal-2019-psycholinguistics}. Forgetting has also been credited as the major limitation for incremental learning models \cite{lee2017overcoming}, \cite{hayes2020remind}, \cite{shi2021overcoming}.
    
    Catastrophic forgetting is being combated in many different ways. However, class ordering is very rarely examined as a means to alleviate the effects of catastrophic forgetting. Class ordering is the idea that the way class data are arranged and fed to an incremental classifier can affect the performance of the model being trained. Class ordering in relation to synthetic neural networks has been examined since the late 80s. Notions of ordering sensitivity have been introduced as a measure of the importance of the order in which a network is fed classes. While class ordering has been examined in class incremental learning \cite{he2022rethinking, masana2020class}, past work seems to only examine vision tasks.
    
    To better understand class ordering within the NLP domain, this paper examines different methods of priming. Semantic, associative and repetition priming methods for NLP systems are inspired by the psychological practice of priming human brains. While current methods of class ordering utilize information contained in confusion matrices \cite{masana2020class}, these require a model to first be trained on the classes to determine which classes the model most often mis-classifies. 
    
    The method that this paper proposes examines class data before training to determine an optimal ordering that limits the effects of catastrophic forgetting. This method is likely to perform better in an unknown environment if the class ordering can be modified to accommodate new classes. Examining semantic and associative relatedness between classes allows us to generate orders for classes that bank on artificial neural networks behaving analogously to the organic phenomenon of priming. 
    
    The contributions of this paper are:
    \begin{itemize}
        \item Explore generalizable class ordering methods that can outperform random ordering for NLP incremental classification.
        \item Identify a class ordering method that can order data before model training occurs,
        \item Draw a connection between biological neural networks and artificial neural networks in the context of priming.
        \item Test the priming methods on different data sets and show that the incremental classification results are better.
    \end{itemize}
    We begin by reviewing past scholarship on open-set classification, class ordering, and priming in order to set the foundation that our research builds off of. We then identify how our research fills a much needed hole in the context of lifelong learning models regarding NLP tasks. Three methods of class ordering are then identified and tested on three datasets across three different incremental learning approaches. From the results we are able to claim that a generalizable class ordering method created before model training occurs does outperform random ordering. As the method stems from research regarding human priming we draw a connection between the two but do not make any empirical claims on the matter as that falls outside the scope of this paper. 
    
    \section{Related Work}
    This work examines the relationship between priming as it relates to humans and how priming can be applied to NLP incremental classification. This relationship is examined in order to identify an optimal class ordering method.
    
    \subsection{Open-set classification}
    Most classification problems in machine learning are evaluated in the ``closed-set" paradigm. This is the scenario in which the set of classes used in training is the same as the set of classes used in testing. This represents the real world inaccurately and begs for the more realistic scenario of ``open-set" classification. Open-set classification is the process of training a model on a set of classes that is less extensive than the set of tested classes \cite{6365193}. Open-set classification has been implemented using loss functions that increase the entropy of softmax scores to better handle background and unknown inputs \cite{dhamija2018reducing}. Others have replaced the softmax layer with the ``Openmax" \cite{Bendale_2016_CVPR} layer that computes how far a piece of data is from known training data based on the penultimate layer's outputs  \cite{prakhya2017open}. 
    
    \subsection{Class ordering}
    Class ordering during classification has been explored since the 1980s \cite{lee1988effects}. However, the majority of efforts in class ordering have only examined computer vision \cite{9615907}. Efforts at ordering typically rely on a confusion matrix that results from a network already training on the entire dataset. To the knowledge of the authors, ours is the first paper to examine class ordering for NLP tasks, as well as class ordering that can be implemented prior to any model training in the context of incremental class learning. As a final claim, we believe this paper is the first to examine the close relationship that biological and synthetic networks have regarding priming. 
    
    \subsection{Priming}
    Priming is the psychological process by which one exposure to a certain stimulus has an influence on the reaction to the exposure of another stimulus  \cite{bargh2014mind}. Psychologists have discovered three types of priming that the authors have identified as viable to be transferred to the incremental learning domain. \emph{Semantic priming} \cite{shelton1992semantic} refers to priming in which the initial stimulus has an important semantic relationship to the reacted stimulus, which influences the reaction to a greater degree than some other stimulus that had a weaker semantic relationship. \emph{Associative priming} \cite{ferrand2004semantic} refers to priming in which the initial stimulus is in close proximity to the reacted stimulus (and hence associated), which influences the reaction to a greater degree than some other stimulus that appears further away in proximity. The final priming method is \emph{repetition priming} \cite{forster1984repetition}. This is priming in which the initial stimulus affects future reactions to itself. As we are examining incremental learning in NLP, these three methods stood out to us. Being able to compute words that have semantic and associative relationships to a given word enables us to see if artificial neural networks are affected similarly to biological neural networks. Our model relies on rehearsal strategies \cite{luo_appraisal_2020} where past data are used for retraining when a new class is discovered. This incremental learning method has deep roots in repetition priming as both are a form of replay of data.

    \begin{figure*}[h!]
      \includegraphics[width=\linewidth]{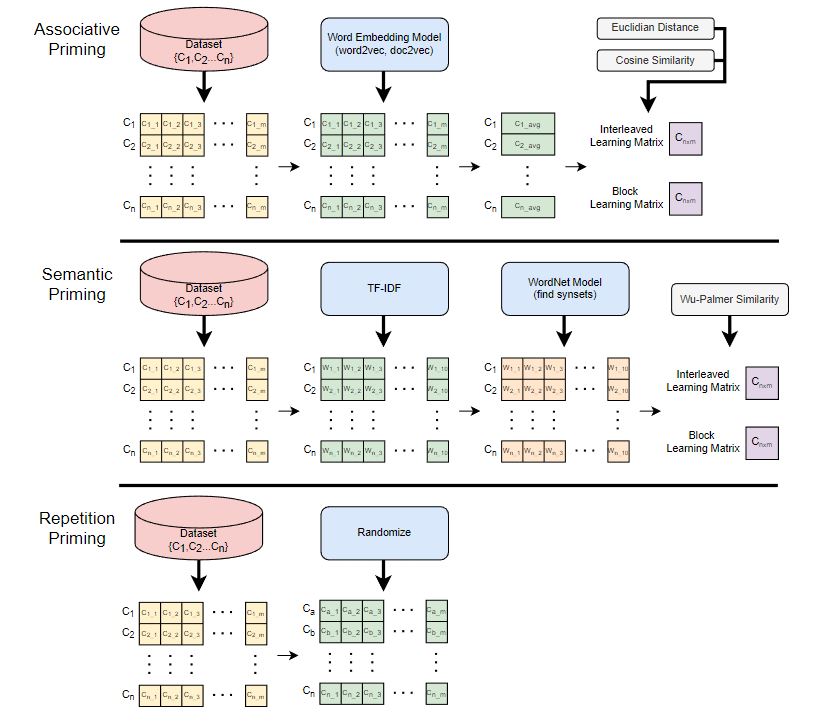}
      \caption{Diagrams showing implementations of the priming techniques used including Associative, Semantic, and Repetition Priming.}
      \label{fig:Priming_Diagrams}
    \end{figure*}
    
    \section{Problem Statement}
    While lifelong machine learning has had some success in computer vision, the same cannot be said for NLP. Lifelong learning models still suffer from catastrophic forgetting when faced with NLP tasks. Class order has been examined for lifelong learning in the context of computer vision, but has been completely untouched in NLP. This paper identifies class ordering methods that outperform random class ordering, which increases accuracy of the model, and hence hampers catastrophic forgetting.
    
    \section{Approach}
    The goal of this paper is to identify class ordering methods that are likely to outperform random class ordering for NLP classification tasks. Three different methods are tested with the aim of producing greater incremental classification accuracy. We also test whether interleaved or block learning \cite{he2022rethinking} performs better in conjunction with the three different methods. The ordering of classes will especially be useful for incremental learning techniques that utilize one main learning model that gains knowledge through a series of iterations where the model is exposed to new data along with old data. This allows the model to maintain class division and minimize catastrophic forgetting. As a result, the incremental approaches analyzed in this paper are ones that focus on data rehearsal.
    The main incremental learning approach used to test the class ordering methods is the Classification Confidence Threshold (CCT) approach \cite{leo_incremental_2021}. The CCT method identifies new classes using open-set classification with a confidence threshold. The learning model used is ``primed" to more easily gain new knowledge, this is shown in Figure \ref{fig:CCT_Model}. We choose this as our primary approach to study because this method leverages the concept of priming in incremental classification and we can modify the approach for deeper exploration of priming techniques. Adding controlled class ordering to this approach would help the model better perform the open-set classification needed and minimize known class confusion. After some optimal class ordering strategies are identified, we also test these approaches on some other rehearsal focused incremental learning methods. The first is the iCaRL \cite{rebuffi2017icarl} approach that learns data from a sequential data stream by producing strong data representations. The second is the EEIL \cite{castro2018end} approach that calculates a learning loss based on a distillation measure to retain old knowledge. Both of these approaches can easily integrate the priming process as it is a data pre-processing step before being used in their respective architectures.

    \subsection{Associative Priming}
    The first heuristic we examined is based on associative relationships between classes. This heuristic is pursued due the effects of associative priming on human learning. To find the associative relationships between classes, an exemplar of the class is first identified by computing the average document for a class. The average document is computed by taking the average of all word embeddings of all documents in a class. While this is a relatively näive method of making an exemplar for a text class, it is computationally cheap and can be computed without the training of anything but a word2vec model \cite{mikolov2013efficient} for the corpus. Word2vec vectors area also trained based on the co-occurances of words in a corpus and this helps capture associative relationships. Along with using word2vec, doc2vec was also used to determine an exemplar by computing the average document as the average of doc2vec representations \cite{le2014distributed}. However, since most of the documents in the selected datasets are short, our approach is practical and reasonably sound.
    
    After an exemplar is computed for each class, the similarity is computed by using cosine similarity. The similarity or dissimilarity between pairs of classes is then stored in an adjacency matrix. The distance between class averages is also computed using the Euclidean distance between them, as an alternative. While cosine similarity tells us how similar classes are to one another, Euclidean distance informs us of how different classes are and where there is little difference, similarity is assumed.
    
    Once an adjacency matrix is computed, an order is constructed based on \emph{interleaved learning}, the process of learning data classes that differ significantly, or \emph{block learning}, the process of learning data classes that are very similar \cite{he2022rethinking}. For interleaved learning, the order is constructed by taking the pair of classes with the greatest distance (or least similarity) as the first two classes and adding on a class to the order that is furthest (least similar) from the last class added. This results in an order of classes that vary the most from each other on an associative level. For block learning, a similar method is used, but instead of using classes that have the greatest distance (lowest similarity), we take classes that have the least distance or most similarity. 
    
    While this method can be made more complex by identifying more robust methods of computing association between classes, this current näive approach offers a starting point for future research in this area. 
    
    \subsection{Semantic Priming}
    The second heuristic we examined is based on semantic relationships between classes. Semantic priming allows for a better learning response when presented with an important representation of a class. This has been shown to be successful with human priming systems \cite{luchkina2021semantic}. To apply semantic priming for incremental learning, we first found a set of the 10 most important words per class using tf-idf \cite{ramos2003using}. For each set of words, the semantic similarity was computed by obtaining a Wordnet synset \cite{fellbaum2010wordnet} of each word and computing the Wu-Palmer similarity \cite{wu1994verb} between them. The Wu-Palmer similarity measures the relatedness of two synsets; this allows us to find the similarity distance between two different sets. Once the average similarity between the words of the two sets is computed, this is then used as the semantic similarity between the two classes. A Wordnet synset contains words and phrases that are fairly synonymous, and hence we consider a synset to be a good semantic representation for words. Calculating the semantic similarity for all the classes results in a similarity adjacency matrix; this is then used for the class ordering to perform either block learning or interleaved learning.
    
    \subsection{Repetition Priming}
    The third heuristic tested is the repetition of the classes. As mentioned earlier, the incremental techniques studied are all rehearsal based methods which involve training and retraining a classifier based on a growing set of classes. Both Associative and Semantic priming generate an ordered list of classes to help reduce catastrophic forgetting; however, a non-primed set of data would have no class ordering restrictions. To test this, repetition priming randomizes the class ordering for each experiment. This process also acts as an ablation study on the effectiveness of priming because the average accuracy from random class orderings can be compared to the average accuracy of the Associative and Semantic priming approaches. To keep the Repetition priming tests similar to the other tests, the class training order for the initial pre-training step is kept the same. A summary of all the priming approaches are shown in Figure 2.
    
    \subsection{Data}
    Two text datasets were used to determine if a method of class ordering could be generally applied across incremental learning NLP tasks. The Reddit Mental Health Dataset (RMHD) consists of posts from 27 SubReddits \cite{low_natural_2020}. The CCAT-50 dataset consists of 50 writing pieces from 50 authors \cite{houvardas2006n}. These two datasets differ greatly in that one consists of internet posts that are filled with slang and vernacular text while CCAT-50 consists of formal writing. Choosing two different styles of NLP datasets helps remove any result bias produced from the word-embedding methods used.

    \section{Results}
    The different priming techniques were first tested on the CCT incremental learning approach. Then, we took the best approach found and applied that to the iCaRL and EEIL approaches.
    In Figures 3 and 4, we present averaged experimental results along with standard deviations for the CCT approach using the RMHD and CCAT-50 datasets, respectively. The figures show the accuracy of the model over 15 iterations using associative priming methods with word2vec. The model starts with training on five classes and with each iteration adds one new class. In both Figures 3 and 4, we see how associative interleaved priming performs much better than associative block priming. This mimics what psychologists have found to be true about humans \cite{rohrer2015interleaved,pan2015interleaving}. Not only is it interesting that the CCT model mimics the same priming effect as humans, but ordering based on associative priming outperforms random ordering by about 15\% for RMHD and 20\% for CCAT-50. The interleaved priming outperforms block priming by 20\% for RHMD and 10\% for CCAT-50. Across both datasets associative interleaved priming outperforms random. It can also be observed that the priming techniques with the best results also produce the least standard deviation and have the most consistent results.
    
    In Figures 5 and 6, we present average experimental data for semantic priming for our datasets. While semantic block priming outperforms interleaved priming for RMHD, this trend is not supported by CCAT-50. The discrepancies arising in the semantic priming results can be explained by the shortcomings of Wordnet. While Wordnet usually produces good results, words not found in the trained vocabulary do not allow for a synset to be created; this is a major problem as most unique word features are not in the vocabulary.
    When examining the most important words for both datasets, many sets of top words included slang, proper nouns, and acronyms, all of which were overlooked by Wordnet. Until semantic similarity can be more easily computed between words that are currently overlooked, this research concludes that semantic priming currently does not provide a good priming solution for all types of text.
    
    For all the following figures, we show just the average results for both the iCaRL and EEIL approaches. In Figures 7-10, we show the results for both datasets for the iCaRL approach. The results obtained are much more consistent than the CCT method, and we once again observe that the associative interleaved priming produces better results than random ordering. This also holds true for the EEIL approach results that are shown in Figures 11-14.

    \begin{figure}[h!]
      \includegraphics[width=\linewidth]{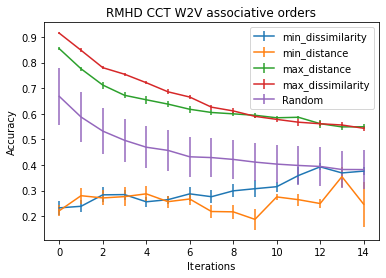}
      \caption{Averaged Associative Priming results for RMHD using CCT. Standard deviations are also shown.}
      \label{fig:boat1}
    \end{figure}
    \begin{figure}[h!]
      \includegraphics[width=\linewidth]{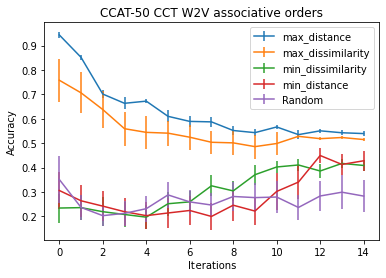}
      \caption{Averaged Associative Priming results for CCAT-50 using CCT. Standard deviations are also shown.}
      \label{fig:boat1}
    \end{figure}
    \begin{figure}[h!]
      \includegraphics[width=\linewidth]{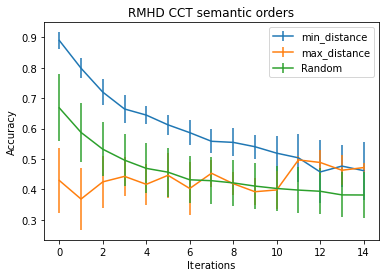}
      \caption{Averaged Semantic Priming results for RMHD dataset using CCT. Standard deviations are also shown.}
      \label{fig:boat1}
    \end{figure}
    \begin{figure}[h!]
      \includegraphics[width=\linewidth]{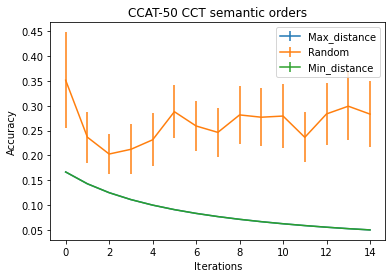}
      \caption{Averaged Semantic Priming results for CCAT-50 using CCT. Standard deviations are also shown.}
      \label{fig:boat1}
    \end{figure}
    
    \begin{figure}[h!]
      \includegraphics[width=\linewidth]{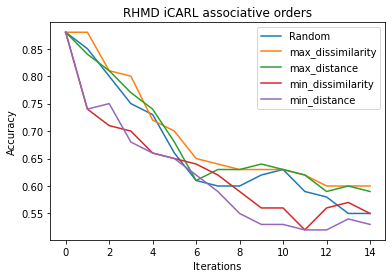}
      \caption{Averaged Associative Priming results for RMHD using iCaRL.}
      \label{fig:boat1}
    \end{figure}
    \begin{figure}[h!]
      \includegraphics[width=\linewidth]{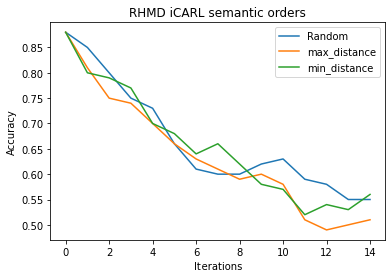}
      \caption{Averaged Semantic Priming results for RMHD dataset using iCaRL.}
      \label{fig:boat1}
    \end{figure}
    \begin{figure}[h!]
      \includegraphics[width=\linewidth]{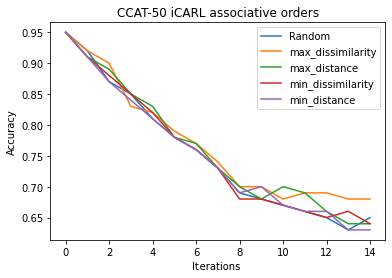}
      \caption{Averaged Associative Priming results for CCAT-50 using iCaRL.}
      \label{fig:boat1}
    \end{figure}
    \begin{figure}[h!]
      \includegraphics[width=\linewidth]{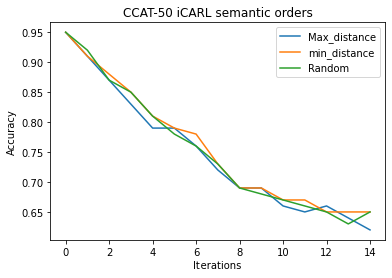}
      \caption{Averaged Semantic Priming results for CCAT-50 using iCaRL.}
      \label{fig:boat1}
    \end{figure}
    
    \begin{figure}[h!]
      \includegraphics[width=\linewidth]{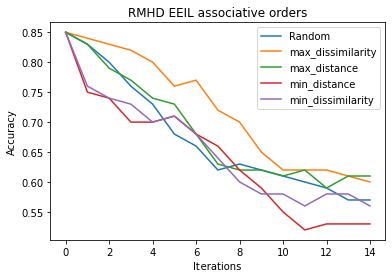}
      \caption{Averaged Associative Priming results for RMHD using EEIL.}
      \label{fig:boat1}
    \end{figure}
    \begin{figure}[h!]
      \includegraphics[width=\linewidth]{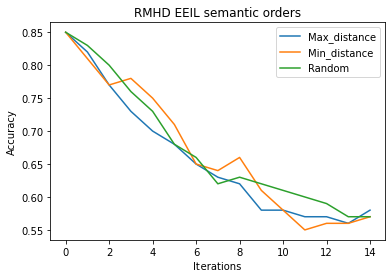}
      \caption{Averaged Semantic Priming results for RMHD using EEIL.}
      \label{fig:boat1}
    \end{figure}
    \begin{figure}[h!]
      \includegraphics[width=\linewidth]{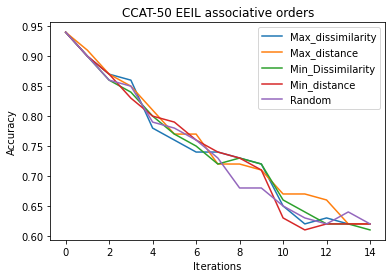}
      \caption{Averaged Associative Priming results for CCAT-50 using EEIL.}
      \label{fig:boat1}
    \end{figure}
    \begin{figure}[h!]
      \includegraphics[width=\linewidth]{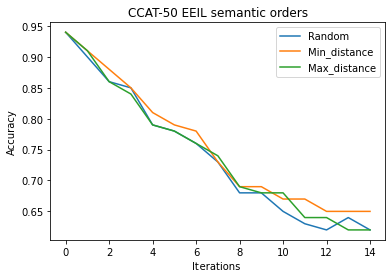}
      \caption{Averaged Semantic Priming results for CCAT-50 using EEIL.}
      \label{fig:boat1}
    \end{figure}

    \section{Analysis and Conclusion}
    Class priming does matter for NLP incremental learning tasks. In the process of examining class priming, we have tested multiple methods and found a method that improves performance over random class ordering. In addition, we also have found another method that under-performs random ordering. Class ordering appears to perform best across both datasets when associated interleaved priming is utilized. This proved to be true for both the RMHD and CCAT-50 datasets for all three incremental learning approaches tested.
    
    Associative interleaved priming is the process of ordering the classes such that the incremental learning model is trained using the most different set of classes found in a dataset. This allows the model to be exposed to the most diverse set of features in earlier training steps; this helps reduce catastrophic forgetting as the model is more ready (more primed) to gain new knowledge without big changes to network parameters. This pattern is also observed in humans as people are more capable of readily learning concepts that are similar to ones they are already familiar with \cite{goodyear2005educational}.
    
    Furthermore, this research shows a deeper similarity between biological and synthetic neural networks in the methods best used to train or consolidate knowledge. Reproduction with other datasets will help solidify our claim that this method is truly generalizable. The results obtained also show that greater research is needed on priming for incremental learning; this is especially true for NLP tasks as more levels of similarity can be found in different types of text. This research is also valuable to most other incremental approaches as most methods do not perform any dataset priming pre-processing. We conclude that associative interleaved priming should be further examined across various NLP incremental classification problems. 
   
    While Visual priming was not experimentally examined, a survey of published papers leads us to believe that similar methods are likely work well on images. The utility of interleaving patterns for visual learning in humans has been studied and found beneficial \cite{brunmair2019similarity}. Visual semantic relationships has had much more research focused on it \cite{vyas2020leveraging,li2019visual}. Visual semantic segmentation methods like CFM, DeepLab, and ParseNet \cite{liu2019visual} can be utilized to better identify the semantic trends between image classes that will eventually make up the similarity matrix. It is our untested hypothesis that this would create a better data ordering for visual classification. Visual associative priming has also been shown to have beneficial effects in human learning \cite{schweinberger1995repetition} and hence we would hope this translates to visual incremental learning. While visual semantic relationships have been explore there seems to be a hole in the research surrounding visual associative relationships \cite{hong2015learning}. In order to experimentally determine how well this method fairs against visual semantic priming, a näive method must be identified.

    \section{Acknowledgements}
    The work reported in this paper is supported by the National Science Foundation under Grant No. 2050919. Any opinions, findings and conclusions or recommendations expressed in this work are those of the author(s) and do not necessarily reflect the views of the National Science Foundation.

    \bibliographystyle{IEEEtran}
    \bibliography{bibliography}
\end{document}